\documentclass[sigconf]{acmart}
\AtBeginDocument{%
  }

\setcopyright{acmlicensed}

\copyrightyear{2027}
\acmYear{2027}
\acmDOI{XXXXXXX.XXXXXXX}
\acmConference[SAC'27]{The 42nd ACM/SIGAPP Symposium on Applied Computing}{April 5--9, 2027}{Gwangju, South Korea}
\acmISBN{979-X-XXXX-XXXX-X/27/04}

\usepackage{float}

\setcopyright{none}
\renewcommand\footnotetextcopyrightpermission[1]{}
\begin{document}


\title{CAMEL: Cross-Architecture Masked Embedding Learning for Ultra-Lightweight Retinal Foundation Models}

\renewcommand{\shorttitle}{CAMEL}

\author{Andre Guarnier De Mitri}
\email{andremitri@usp.br}
\authornotemark[1]
\orcid{0009-0006-2197-3460}
\author{Giovani Decico Lucafo}
\email{giovanilucafo@usp.br}
\affiliation{%
  \institution{ICMC, University of São Paulo}
  \city{São Carlos}
  \state{São Paulo}
  \country{Brazil}
}

\author{Miller Matheus Lima Anacleto Rocha}
\email{anacletomiller@usp.br}
\author{Diego Furtado Silva}
\email{diegofsilva@usp.br}
\affiliation{%
  \institution{ICMC, University of São Paulo}
  \city{São Carlos}
  \state{São Paulo}
  \country{Brazil}
}


\renewcommand{\shortauthors}{A.G. De Mitri et al.}
\settopmatter{printacmref=false}
\setcopyright{none}
\renewcommand\footnotetextcopyrightpermission[1]{}
\pagestyle{plain}

\begin{abstract}

Recent progress in deep learning has substantially advanced automated grading of ophthalmologic pathologies, supported by large annotated datasets and improved representation learning. However, a significant gap remains between state-of-the-art performance and clinical scalability. In particular, foundation models have reshaped image-based medical screening and diagnosis. These large-scale architectures trained on extensive datasets learn rich, generalizable feature representations that can be adapted to a wide range of downstream tasks. In ophthalmology, RETFound, one of the most relevant foundation models for retinal images, performs strongly across classification, segmentation, and risk prediction tasks. Despite their effectiveness, foundation models are often computationally prohibitive, hindering their deployment in real-world clinical environments where hardware constraints and latency are critical. To bridge this gap, we introduce CAMEL (Cross-Architecture Masked Embedding Learning), a novel cross-architecture distillation framework that uses an asymmetric masking strategy during embedding learning to effectively transfer knowledge from massive teacher foundation models to compact, lightweight student models. Our experimental evaluation across four datasets shows that CAMEL preserves accuracy while drastically reducing computational complexity, enabling lightweight yet highly accurate models suitable for resource-limited clinical settings.

\end{abstract}



\maketitle

\section{Introduction}
Recent progress in deep learning has substantially advanced automated grading of ophthalmologic pathologies, supported by large annotated datasets and improved representation learning~\cite{sevgi2024foundation}. However, a significant gap remains between state-of-the-art performance and clinical scalability~\cite{TING2019100759}. In particular, foundation models have reshaped the medical imaging landscape. These large-scale architectures trained on extensive datasets learn rich, generalizable feature representations that easily adapt to a wide range of downstream tasks~\cite{bommasani2022}. In ophthalmology, RETFound~\cite{zhou2023RETFound}, a prominent foundation model trained on vast retinal image repositories, has demonstrated exceptional performance across classification, segmentation, and risk prediction tasks.

Despite their effectiveness, foundation models are computationally restrictive, severely hindering their deployment in real-world clinical environments where hardware constraints, energy consumption, and latency are critical. Model distillation~\cite{hinton2015distillation} has emerged as a promising strategy for transferring knowledge from a massive teacher model to a compact student model. This approach is particularly valuable in medical applications, enabling real-time screening on diverse instruments, including mobile devices~\cite{mbrset}. However, conventional distillation approaches often struggle to bridge structurally different architectures.

To address these limitations, we introduce CAMEL (Cross-Archi\-tecture Masked Embedding Learning), a novel feature-distillation framework that transfers representations across heterogeneous architectures using masked distillation. To demonstrate the versatility of CAMEL, we evaluate its performance across a broad spectrum of lightweight student backbones, including Convolutional Neural Networks (\textit{MobileNet}~\cite{mobilenet}, \textit{EfficientNet}~\cite{EfficientNet}, and a standard \textit{U-Net}~\cite{unet} encoder), a Vision Transformer (\textit{TinyViT}~\cite{tinyvit}), and a State-Space Model (\textit{Vision Mamba - Vim}~\cite{zhu2024Vim}).

\begin{figure}[t]
    \centering
    \includegraphics[width=1.0\linewidth]{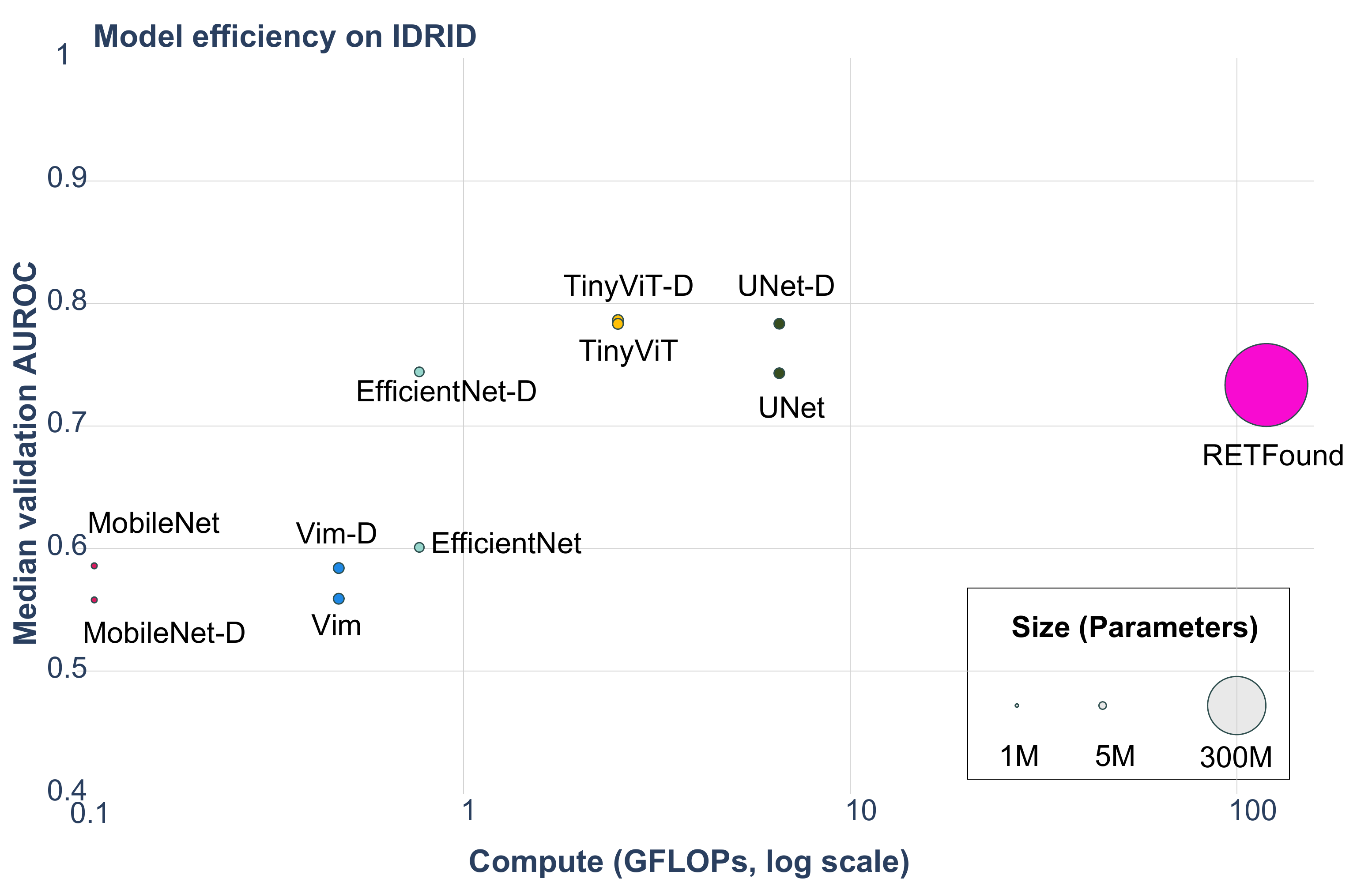}
    \caption{Comparison of parameter count, computational cost, and AUROC between RETFound, a large Transformer-based foundation model, and a series of lightweight models for the dataset IDRID. The suffix ``D'' represents models distilled from RETFound.}
    \label{fig:VMamba_RETFound_complexity}
\end{figure}

To assess model performance, we evaluate four publicly available benchmark datasets for Diabetic Retinopathy (DR) grading: IDRiD~\cite{porwal2018IDRiD}, APTOS 2019~\cite{kassani2019APTOS}, mBRSET~\cite{mbrset}, and MESSIDOR-2~\cite{Decenciere2014messidor}. Including mBRSET is particularly meaningful, as its images are captured via mobile devices, reflecting the unconstrained, low-resource clinical conditions our lightweight models target. Together, these benchmarks present significant heterogeneity in image quality, illumination, hardware, and patient demographics, serving as a rigorous testbed for cross-domain generalization.

As illustrated in Fig.~\ref{fig:VMamba_RETFound_complexity}, our empirical results show that distilled student models consistently outperform their counterparts trained from scratch across all evaluated architectures. Furthermore, we show that masked feature distillation~\cite{yang2022maskedgenerativedistillation} provides greater robustness under domain shifts than unmasked distillation. By benchmarking a wide range of architectures within CAMEL, we offer a comprehensive spectrum of operational choices, enabling practitioners to select the optimal trade-off between computational cost and diagnostic precision according to their deployment constraints. Finally, our code and implementation details are publicly available\footnote{\url{https://anonymous.4open.science/r/CAMELL/}}.

In summary, our main contributions are threefold:
\begin{itemize}
    \item We propose the Cross-Architecture Masked Embedding Learning (CAMEL) framework, a novel feature-distillation framework that leverages asymmetric masked embedding alignment to transfer knowledge from large-scale foundation models (such as RETFound) into compact, lightweight student backbones across heterogeneous architectural paradigms.
    
    \item We comprehensively evaluate CAMEL across diverse vision backbones, demonstrating that distilled models usually outperform their counterparts trained from scratch and providing a flexible spectrum of accuracy-efficiency trade-offs.

    \item We make our complete implementation codebase, model configurations, and pre-processing pipeline publicly available to foster further research into resource-constrained medical imaging deployment.
\end{itemize}


\section{Related Work}\label{sec:related}

The transition from high-performance research models to practical clinical tools requires navigating several technical bottlenecks. This section reviews three complementary research directions that motivate our approach: general self-supervised learning and foundation models, their adaptation to ophthalmology, and RETFound, a specific, recent, yet widely used foundation model for retinal images.


\subsection{Foundation Models and Self-Supervised Learning}
Foundation models have fundamentally influenced modern machine learning by acting as versatile ``base'' systems that learn general-purpose features. As defined by Bommasani et al.~\cite{bommasani2022}, these models are trained on massive datasets and can be adapted to a wide variety of downstream tasks. Their success is largely driven by self-supervised learning (SSL), which enables networks to learn deep semantic representations without exhaustive human labeling.

Among SSL methods, DINO~\cite{caron2021dino} demonstrated that Vision Transformers (ViTs) could learn object-centric representations through self-supervision by having a ``student'' network predict the outputs of a ``teacher'' model. This teacher-student paradigm is central to our proposal; however, while DINO typically employs homogeneous architectures, we explore a heterogeneous distillation strategy, transferring knowledge from a massive Transformer-based teacher to a State Space Model-based student or Convolutional Neural Networks (CNNs).

\subsection{Foundation Models in Ophthalmology}

\begin{figure*}[t]
    \centering
    \includegraphics[width=\linewidth]{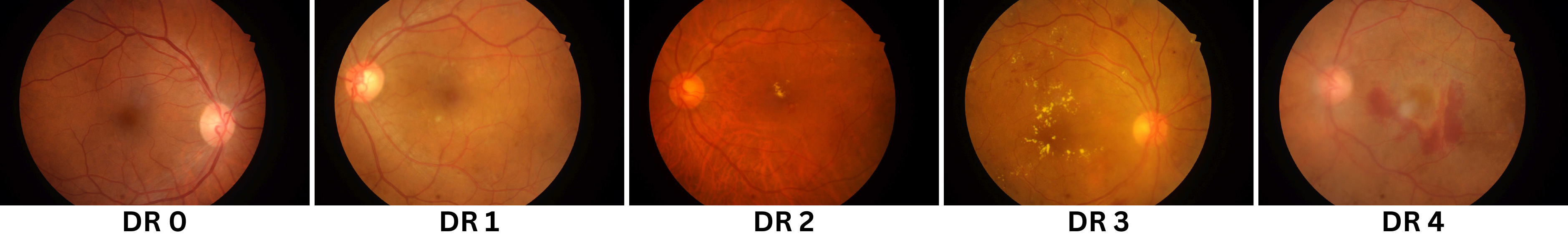}
    \caption{Classification task, IDRiD sample images illustrating diabetic retinopathy severity levels (DR0–DR4). Diabetic retinopathy is a diabetes complication that progressively damages retinal blood vessels and can lead to vision loss}
    \label{fig:DR}
\end{figure*}

Inspired by their success on natural images, foundation models have been specialized for several application domains, including retinal imaging (e.g., Figure~\ref{fig:DR}), where they are designed to handle the distinctive textures and structures of the eye. The most significant milestone in this domain is RETFound, introduced by Zhou et al.~\cite{zhou2023RETFound}. Pretraining a large Vision Transformer (ViT) on 1.6 million unlabeled fundus images using Masked Autoencoder (MAE), RETFound established a new state-of-the-art for ophthalmic tasks, acting as a robust feature extractor from retinal images. However, this power comes at a high computational price: 303M parameters and 119.29 GFLOPs per forward pass.

Other ophthalmic foundation models explore different modalities and training strategies. RetCLIP~\cite{du2024retclip} aligns retinal images with clinical text reports through contrastive language, enabling zero-shot inference and cross-modal retrieval. EyeFounder~\cite{shi2024eyefound} further extends foundation models to broader ocular datasets and three-dimensional imaging modalities. Despite their expanded scope, such models remain computationally expensive, motivating the need for efficient distilled alternatives.

\section{CAMEL: Cross-Architecture Masked Embedding Learning} \label{sec:lemur}

The primary challenge in deploying models like RETFound is the Transformer bottleneck. Although Transformers dominate current benchmarks~\cite{zhou2023RETFound,du2024retclip,shi2024eyefound}, their quadratic complexity~\cite{heidari2024efficiency} in sequence length limits scalability for high-resolution medical images and their use in resource-constrained environments. To address this, we move from heavy, attention-based foundation models to highly efficient, lightweight architectures.



One strategy to achieve this computational reduction is transferring the representation capabilities of the large foundation model into diverse lightweight student architectures. Rather than being restricted to homogeneous distillation, our framework supports a wide spectrum of compact backbones, from structurally similar, lightweight Vision Transformers (such as TinyViT~\cite{tinyvit}) to heterogeneous paradigms including Convolutional Neural Networks (CNNs) and selective State-Space Models (SSMs, such as Vision Mamba - Vim~\cite{zhu2024Vim}). This flexible cross-architecture alignment enables student models to inherit the teacher’s spatial and semantic representations while leveraging their respective architectural efficiencies for linear-time inference or reduced operational footprints.

\begin{figure}[!htb]
\centering
\includegraphics[width=\linewidth]{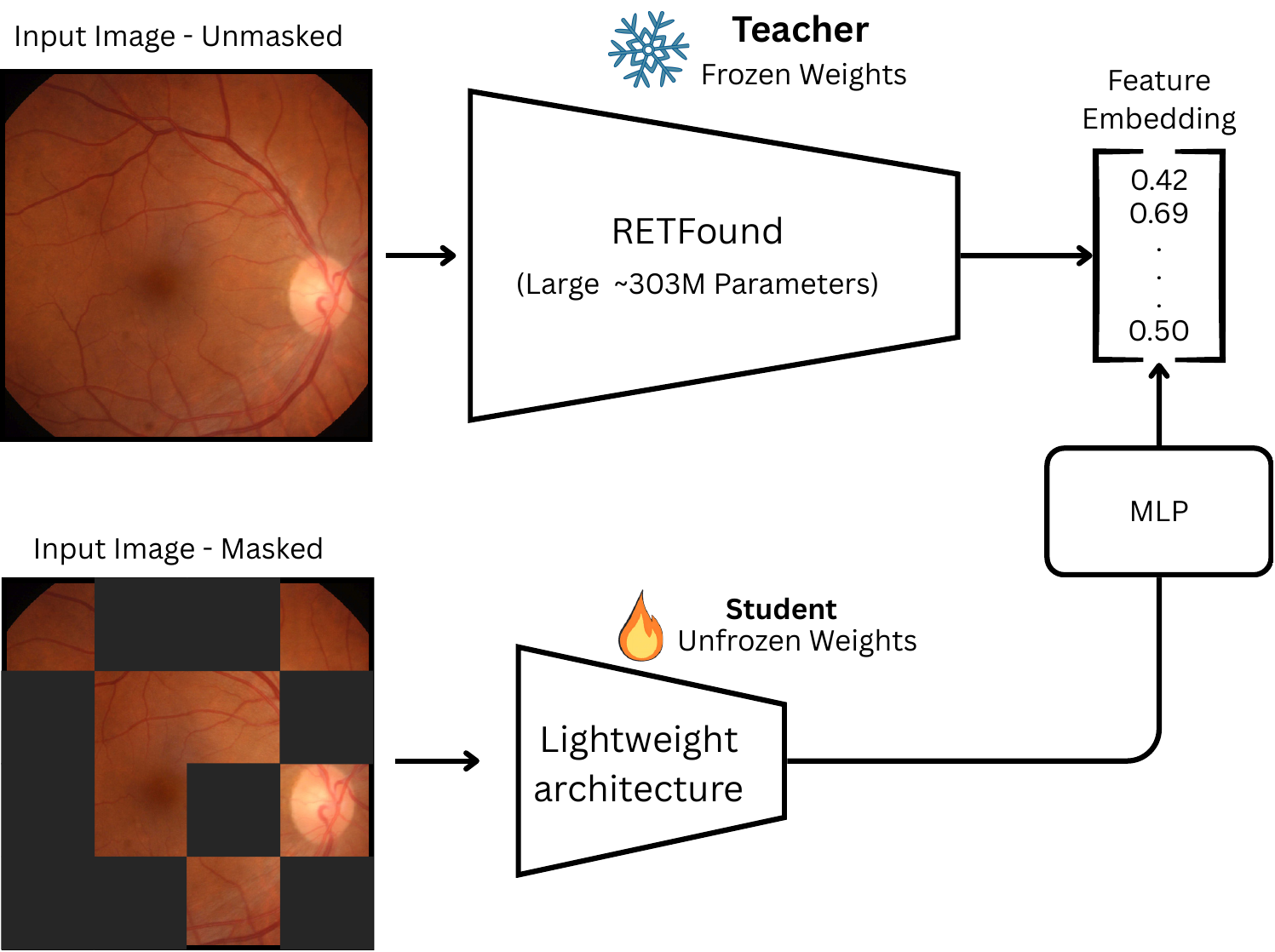}
\caption{CAMEL framework: distillation process for training the student backbone using RETFound as the teacher model. In this configuration, the ViM student processes a masked version of the input image, whereas the teacher receives the full, unmasked image; an MLP projection layer approximates the student's and teacher's feature embeddings. 
}
\label{fig:dist}
\end{figure}

To capture RETFound's representational strengths in a computationally efficient architecture, we propose distilling the knowledge comprised in its feature embeddings into a lightweight model. As illustrated in Figure~\ref{fig:dist}, our approach uses a teacher-student paradigm in which the RETFound teacher remains frozen throughout the process. This ensures that the teacher acts as a static source of high-quality retinal features, while the lightweight student is optimized to minimize the discrepancy between their respective embedding spaces.

To evaluate feature reconstruction under varying spatial constraints, we explore both unmasked and masked distillation strategies. In the unmasked setting, both the student and teacher process identical inputs, using complete fundus images. Conversely, in the masked configuration, we apply an MAE-style random masking strategy~\cite{he2022masked} to the student's input. Specifically, we randomly mask 75\% of the image patches using a uniform distribution without replacement, while the frozen RETFound teacher receives the full, unmasked image. This forces the student model to learn rich spatial dependencies and infer missing structural details by aligning its projected representations with the complete teacher embeddings. Initial empirical evaluations on both the IDRiD and APTOS datasets demonstrated that this masked distillation strategy yields superior feature representations for lightweight student backbones. Consequently, we selected the masked approach as our standard distillation protocol for comparing models.

The distillation objective is to align the student's internal representations with the teacher's feature geometry. This is achieved by minimizing the cosine distance between the projected student features and the teacher's embeddings, defined by the following loss function:


\begin{equation}
\mathcal{L}_{\text{distill}} = \frac{1}{B} \sum_{i=1}^{B} \left(1 - \frac{\mathbf{s}_i^{\text{proj}} \cdot \mathbf{t}_i}{|\mathbf{s}_i^{\text{proj}}| |\mathbf{t}_i|}\right),
\end{equation}

\noindent where $\mathbf{s}_i^{\text{proj}}$ and $\mathbf{t}_i$ denote the projected student features and the corresponding teacher features for the $i$-th sample, respectively, in a batch of size $B$.



After the distillation phase, the student backbone is finetuned to better adapt the distilled retinal knowledge. We then perform a supervised downstream finetuning phase to adapt the model for Diabetic Retinopathy (DR) grading. For this purpose, we attach a classification head inspired by the DINO framework~\cite{caron2021dino}, consisting of two fully connected layers with Batch Normalization, GELU activations, and dropout, followed by a final linear layer that produces the class logits. This two-stage pipeline allows CAMEL to achieve foundation-level diagnostic accuracy while remaining light enough for real-time inference on mobile devices.

The remainder of this section details the backbone architecture illustrated in Fig.~\ref{fig:dist}.

\subsection{Projection layer}
To align the representation spaces of the teacher and student models, we introduce a lightweight projection module that maps the student embedding dimension to the teacher's feature dimension.  
The projector is implemented as a two-layer MLP with a GELU activation, consisting of a first linear layer with 128 neurons (expanding the 64-dimensional student embedding) followed by a second linear layer with 1024 neurons corresponding to the teacher embedding dimension. This component contains 468K parameters, adding negligible computational overhead while proving essential for stable distillation.

\subsection{Optimizer}
Based on the optimization strategy used in RETFound, we adopt the AdamW optimizer for both the distillation stage and the subsequent downstream task. As our experiments confirm (cf. Section~\ref{sec:results}), this choice provides stable convergence and effective regularization, which are particularly important for models derived from sequence-processing architectures such as Vim. Training stability is further enhanced by a learning rate schedule that combines a warmup phase with cosine annealing: the learning rate increases linearly over the first 10 epochs and then decays smoothly toward a small final value as training progresses.

Across all experiments, we maintain a consistent configuration of hyperparameters. The base learning rate is set to $5 \times 10^{-4}$, gradually reduced to a final value of $1 \times 10^{-6}$ by the end of training. Weight decay is fixed at $1 \times 10^{-4}$ to provide additional regularization throughout optimization.

\subsection{Metrics and Callbacks}\label{sec:metrics_callbacks}
During the distillation phase, the objective is to align the features between the RETFound teacher and the lightweight student.  
Thus, we monitor the distillation loss, computed as the cosine distance between the projected student features and the teacher features.  

To ensure stable convergence and avoid overfitting to noisy feature alignments, we employ two key callbacks. The first one is Early Stopping, triggered when the validation distillation loss stops improving for 15 consecutive epochs. We also employ Model Checkpointing, saving only the checkpoint with the lowest validation distillation loss.

After distillation, a second training phase is performed where the distilled backbone and a lightweight classification head are trained using supervised labels. In this stage, different metrics are tracked, such as accuracy, macro-F1, AUROC, and AUPR, together with early stopping and checkpointing based on the validation F1 score, ensuring optimal downstream classification performance.

\subsection{Image Preprocessing and Data Augmentation}
In order to improve feature extraction and mitigate inherent variations in fundus image quality (e.g., uneven illumination and camera artifacts), we apply a preprocessing pipeline prior to network ingestion. 

First, extraneous black backgrounds are removed using contour-based bounding box cropping. We then isolate the green color channel, which empirically yields the highest contrast for critical retinal structures such as blood vessels and lesions. To correct uneven illumination across the fundus, we apply Ben Graham's luminance normalization method, combining the original green channel with a Gaussian-blurred version to equalize lighting. This is followed by Contrast Limited Adaptive Histogram Equalization (CLAHE)~\cite{PIZER1987355} to enhance local structural details. Finally, a central circular mask is applied to eliminate artificial corner artifacts, and the enhanced grayscale image is converted back to a 3-channel RGB format for compatibility with the pretrained feature spaces.

During the downstream fine-tuning phase, we employ an augmentation strategy to improve generalization. This regime begins with standard geometric variations, including random horizontal and vertical flips and random rotations up to $45^{\circ}$. To make the model anatomically robust against variations in eye positioning and lens geometry, we incorporate elastic deformations ($\alpha=50.0$). Finally, we apply RandAugment~\cite{9150790} to optimize the augmentation policy space, then apply standard ImageNet normalization before passing the tensors to the network.

\section{Experimental Setup} \label{sec:experiments}

We evaluate the CAMEL framework using four retinal fundus datasets: IDRiD, APTOS, mBRSET, and MESSIDOR-2. These datasets vary substantially in size, image quality, acquisition hardware, and patient demographics, providing an environment to assess the model's adaptability and performance consistency.

To build a comprehensive benchmark, we compare it against a variety of vision models. Our lightweight convolutional baselines include MobileNet~\cite{mobilenet}, EfficientNet-B0~\cite{EfficientNet}, and a standard U-Net encoder~\cite{unet}. For Vision Transformers, we employ TinyViT-5M-224~\cite{tinyvit} and, as a State Space Model, we use Vim~\cite{zhu2024Vim}. To assess the impact of our approach, we train each student backbone from scratch as a baseline and compare it directly with its distilled counterpart. This paired configuration allows us to isolate the performance gains from knowledge distillation within each evaluated architectural family. Finally, we benchmark all compact models against the full-scale RETFound teacher to establish a speculative performance upper bound and measure the diagnostic capacity retained post-distillation. Table~\ref {tab:model_complexity} details the computational complexity and parameter counts for all evaluated architectures, as well as estimated energy and carbon dioxide usage.

\begin{table}[!htb]
\centering
\caption{Computational complexity, inference efficiency, and environmental footprint across evaluated base architectures using a NVIDIA RTX A5000 GPU. Latency is measured per image at batch size 1, while throughput represents images processed per second at batch size 32. Energy and CO$_2$ metrics are scaled by $10^{-6}$ for readability. Results are aggregated across all runs (mean $\pm$ std).}
\label{tab:model_complexity}
\setlength{\tabcolsep}{2pt}
\resizebox{\linewidth}{!}{%
\begin{tabular}{lcccccc}
\toprule
\textbf{Model} & \textbf{\begin{tabular}[c]{@{}c@{}}Params\\ (M)\end{tabular}} & \textbf{GFLOPs} & \textbf{\begin{tabular}[c]{@{}c@{}}Latency\\ (ms)\end{tabular}} & \textbf{\begin{tabular}[c]{@{}c@{}}Throughput\\ (img/s)\end{tabular}} & \textbf{\begin{tabular}[c]{@{}c@{}}Energy \\ ($10^{-6}$ kWh)\end{tabular}} & \textbf{\begin{tabular}[c]{@{}c@{}}CO$_2$ \\ ($10^{-6}$ kg)\end{tabular}} \\ \midrule
MobileNet & 1.52  & 0.111  & 1.18 $\pm$ 0.02   & 11611.9 $\pm$ 82.5 & 0.95 $\pm$ 1.95 & 0.09 $\pm$ 0.19 \\
EfficientNet  & 4.01  & 0.769  & 1.83 $\pm$ 0.04   & 2649.6 $\pm$ 3.2   & 8.50 $\pm$ 2.10 & 0.84 $\pm$ 0.21 \\
UNet  & 4.85  & 6.561  & 0.71 $\pm$ 0.00   & 2348.2 $\pm$ 30.9   & 8.60 $\pm$ 1.85 & 0.84 $\pm$ 0.18 \\
Vim   & 5.02  & 0.476  & 17.46 $\pm$ 0.75  & 1691.2 $\pm$ 77.8   & 18.15 $\pm$ 1.70 & 1.78 $\pm$ 0.17 \\
TinyViT  & 5.07  & 2.510  & 1.92 $\pm$ 0.03   & 1695.6 $\pm$ 7.4   & 14.40 $\pm$ 2.20 & 1.41 $\pm$ 0.21 \\ \hline
\multicolumn{7}{l}{\textbf{Foundation Teacher}}      \\
RETFound  & 303.31 & 119.292   & 14.79 $\pm$ 0.12  & 93.2 $\pm$ 0.9 & 278.50 $\pm$ 2.90  & 27.38 $\pm$ 0.28  \\ \bottomrule
\end{tabular}%
}
\end{table}

The Indian Diabetic Retinopathy Image Dataset (IDRiD)~\cite{porwal2018IDRiD} serves as our primary benchmark for both feature distillation and downstream classification. After standardizing the preprocessing and split construction, we obtained training and validation sets composed of 413 and 103 images, respectively. Despite its relatively small size, IDRiD's high-quality annotations make it an essential baseline for evaluating the precision of distilled ophthalmic features.

To examine scalability, we further train and evaluate the models on the APTOS 2019 Blindness Detection dataset~\cite{kassani2019APTOS}. This dataset contains 3,662 labeled retinal images, from which we allocated 2,930 for training and 366 for validation. Its larger scale and distinct distributional characteristics relative to IDRiD let us assess how well models adapt to diverse clinical imaging conditions and varying grading protocols across populations.

We employ the mBRSET~\cite{mbrset} dataset to assess the robustness under realistic acquisition conditions. mBRSET consists of fundus images captured with handheld portable cameras in community screening settings, exhibiting significant variations in illumination, focus, and field-of-view compared to the high-end desktop cameras used in IDRiD and APTOS. Notably, we first evaluate mBRSET without additional fine-tuning to assess the viability of masked and unmasked variants. Afterward, this dataset is fine-tuned for in-domain testing of all architectures. We acknowledge that this dataset is highly imbalanced, with a predominant majority of healthy (DR0) samples and only 1.68\% representing severe (DR3) cases, posing a significant challenge for robust automated grading.

Furthermore, we incorporate the MESSIDOR-2 dataset~\cite{Decenciere2014messidor}, a widely recognized standard for diabetic retinopathy assessment. It comprises 1,748 macula-centered fundus images collected across 874 patient examinations. Including MESSIDOR-2 allows us to validate the framework's learning capacity on an established international benchmark.


To comprehensively evaluate the effectiveness of each approach, we structured our experimental protocol to directly train and evaluate the model independently on all four datasets. In this approach, both feature distillation and downstream classification training are conducted using the respective training split of each individual source dataset, then evaluated on the evaluation split. This enables us to establish robust in-domain performance metrics across diverse clinical settings, ranging from high-end desktop imaging to unconstrained portable community screenings. To ensure the reliability of our findings, all experiments are conducted across five random seeds, and statistical significance is determined using a non-parametric Friedman test followed by a Nemenyi post-hoc analysis to compare the performance of the lightweight models.

\section{Results \& Discussion} \label{sec:results}

In this section, we first analyze the effectiveness of the masked and unmasked variants of the CAMEL framework by testing them on Vim models and comparing them with the RETFound teacher for in-domain and cross-domain evaluation, as well as a Zero-Shot test on mBRSET. Once we establish the best alternative across datasets, we use it in all subsequent experiments for UNet, EfficientNet, Mobilenet, and TinyVIT. We trained and evaluated all models on IDRiD, APTOS, mBRSET, and MESSIDOR.

\subsection{Distillation Masking Analysis}

\begin{figure}[!htb]
    \centering
    \includegraphics[width=\linewidth]{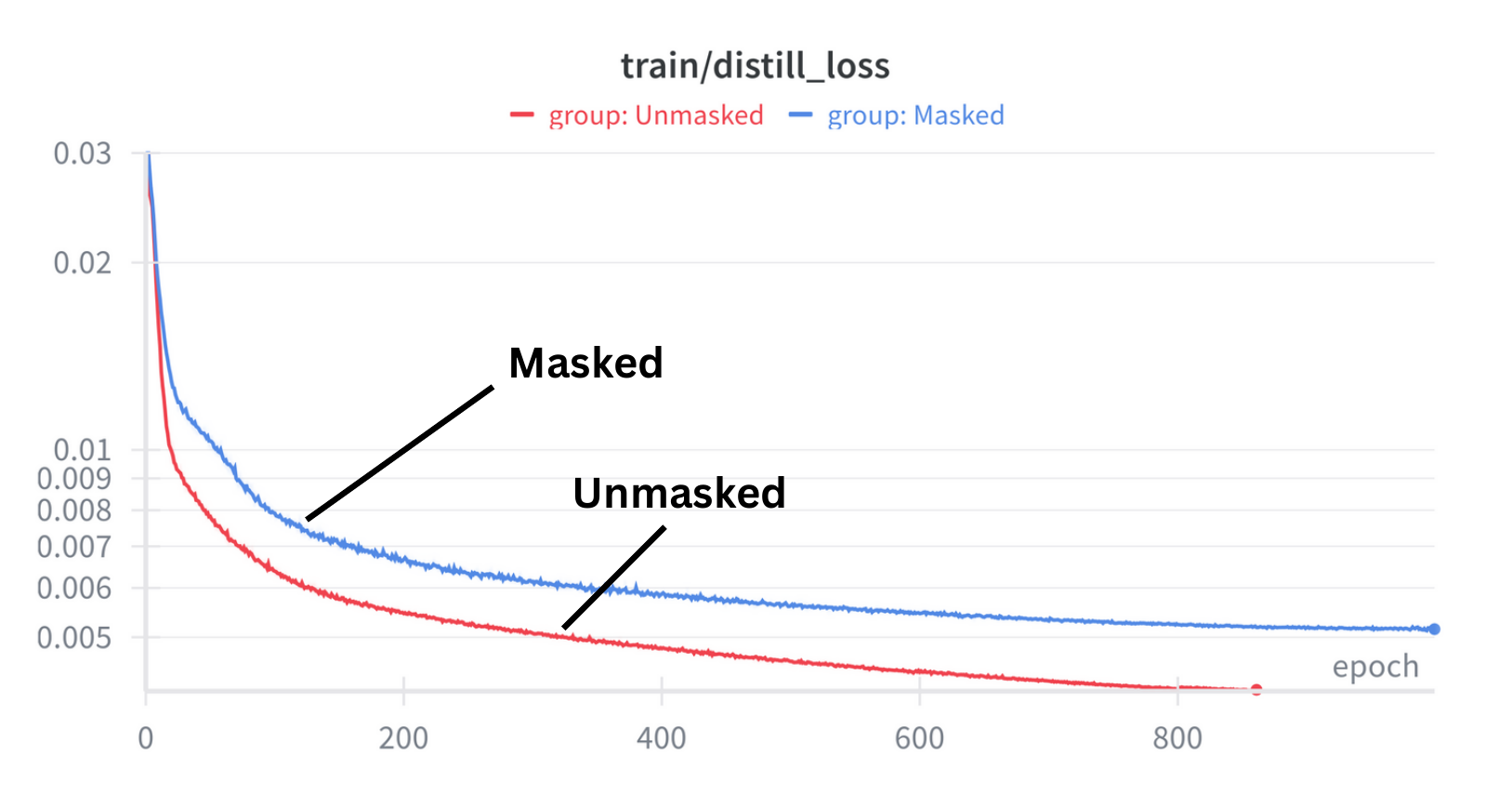}
    \caption{Average loss during the distillation of the RETFound teacher and the masked/unmasked Vim-based students.}
    \label{fig:RETFound_distance}
\end{figure}

To investigate how distillation strategy shapes representation learning, Figure~\ref{fig:RETFound_distance} illustrates the trajectory of the cosine distance loss between the RETFound teacher and student configurations throughout training. For this analysis, we select the Vision Mamba (Vim) backbone as our representative student, as its linear-time state-space mechanisms offer a compelling non-Transformer paradigm to probe how distinct inductive biases reconstruct teacher representations under spatial constraints. While the unmasked student aligns more closely with the teacher's representation, the masked variant maintains a slightly higher yet remarkably stable distance. This behavior indicates that asymmetric masking forces the student model to learn a distinct, structurally robust feature space. From a practical standpoint, this suggests that masking the input acts as a regularizer once the model no longer closely mimics the teacher’s representation.

To determine the best configuration, Table~\ref{tab:consolidated_ablation} consolidates the in-domain and cross-dataset generalization results across the IDRiD, APTOS, and mBRSET datasets. In-domain evaluations demonstrate that while RETFound achieves the highest overall performance, both student variants remain highly competitive. However, the cross-dataset evaluations reveal a critical distinction. Under severe domain shifts (e.g., training on APTOS and evaluating on IDRiD), the masked student demonstrates superior generalization capability, notably outperforming its unmasked counterpart in accuracy and ranking metrics. 


\begin{table}[!htb]
\centering
\caption{Masked vs. unmasked variants across validation and cross-dataset evaluations (mean $\pm$ std). The best results in each column per dataset are \textbf{bolded}, and the second best are \underline{underlined}.}
\label{tab:consolidated_ablation}
\resizebox{\linewidth}{!}{%
\begin{tabular}{lcccc}
\toprule
Model & Acc & F1 & AUROC & AUPR \\
\midrule
\multicolumn{5}{l}{\textbf{In-Domain IDRiD}} \\
Vim (Masked)   & 0.412 $\pm$ 0.033 & 0.318 $\pm$ 0.023 & 0.610 $\pm$ 0.020 & 0.350 $\pm$ 0.008 \\
Vim (Unmasked) & \underline{0.454 $\pm$ 0.021} & \underline{0.330 $\pm$ 0.011} & \underline{0.626 $\pm$ 0.027} & \underline{0.356 $\pm$ 0.020} \\
RETFound (Teacher) & \textbf{0.516 $\pm$ 0.047} & \textbf{0.390 $\pm$ 0.024} & \textbf{0.665 $\pm$ 0.029} & \textbf{0.393 $\pm$ 0.020} \\
\midrule
\multicolumn{5}{l}{\textbf{Cross: Train APTOS $\rightarrow$ Eval IDRiD}} \\
Vim (Masked)   & \underline{0.213 $\pm$ 0.051} & \underline{0.146 $\pm$ 0.029} & \underline{0.554 $\pm$ 0.021} & \underline{0.240 $\pm$ 0.009} \\
Vim (Unmasked) & 0.193 $\pm$ 0.038 & 0.129 $\pm$ 0.016 & 0.530 $\pm$ 0.013 & 0.236 $\pm$ 0.009 \\
RETFound (Teacher) & \textbf{0.317 $\pm$ 0.045} & \textbf{0.255 $\pm$ 0.038} & \textbf{0.664 $\pm$ 0.015} & \textbf{0.319 $\pm$ 0.015} \\
\midrule
\multicolumn{5}{l}{\textbf{In-Domain APTOS}} \\
Vim (Masked)   & 0.750 $\pm$ 0.020 & \underline{0.569 $\pm$ 0.011} & 0.872 $\pm$ 0.019 & \underline{0.574 $\pm$ 0.015} \\
Vim (Unmasked) & \underline{0.762 $\pm$ 0.006} & 0.559 $\pm$ 0.011 & \underline{0.879 $\pm$ 0.005} & 0.571 $\pm$ 0.009 \\
RETFound (Teacher) & \textbf{0.812 $\pm$ 0.010} & \textbf{0.663 $\pm$ 0.021} & \textbf{0.922 $\pm$ 0.007} & \textbf{0.665 $\pm$ 0.015} \\
\midrule
\multicolumn{5}{l}{\textbf{Cross: Train IDRiD $\rightarrow$ Eval APTOS}} \\
Vim (Masked)   & \underline{0.290 $\pm$ 0.132} & \underline{0.173 $\pm$ 0.044} & 0.558 $\pm$ 0.033 & 0.238 $\pm$ 0.025 \\
Vim (Unmasked) & 0.214 $\pm$ 0.114 & 0.148 $\pm$ 0.068 & \underline{0.575 $\pm$ 0.051} & \underline{0.253 $\pm$ 0.035} \\
RETFound (Teacher) & \textbf{0.439 $\pm$ 0.095} & \textbf{0.268 $\pm$ 0.033} & \textbf{0.603 $\pm$ 0.040} & \textbf{0.271 $\pm$ 0.026} \\
\midrule
\multicolumn{5}{l}{\textbf{Zero-Shot: Train APTOS $\rightarrow$ Eval mBRSET}} \\
Vim (Masked)   & 0.590 $\pm$ 0.190 & \underline{0.180 $\pm$ 0.009} & 0.498 $\pm$ 0.016 & 0.201 $\pm$ 0.004 \\
Vim (Unmasked) & \textbf{0.755 $\pm$ 0.016} & \underline{0.180 $\pm$ 0.009} & \underline{0.544 $\pm$ 0.020} & \underline{0.210 $\pm$ 0.005} \\
RETFound (Teacher) & \underline{0.725 $\pm$ 0.028} & \textbf{0.236 $\pm$ 0.025} & \textbf{0.626 $\pm$ 0.017} & \textbf{0.249 $\pm$ 0.006} \\
\midrule
\multicolumn{5}{l}{\textbf{Zero-Shot: Train IDRiD $\rightarrow$ Eval mBRSET}} \\
Vim (Masked)   & \underline{0.637 $\pm$ 0.051} & \underline{0.193 $\pm$ 0.006} & 0.492 $\pm$ 0.022 & \underline{0.204 $\pm$ 0.007} \\
Vim (Unmasked) & 0.486 $\pm$ 0.313 & 0.145 $\pm$ 0.061 & \underline{0.493 $\pm$ 0.025} & \underline{0.204 $\pm$ 0.009} \\
RETFound (Teacher) & \textbf{0.674 $\pm$ 0.096} & \textbf{0.216 $\pm$ 0.016} & \textbf{0.587 $\pm$ 0.035} & \textbf{0.230 $\pm$ 0.013} \\
\bottomrule
\end{tabular}
    }
\label{table:masked_vs_unmasked}
\end{table}

\subsection{Benchmark Evaluations}

\begin{figure*}[!htb]
    \centering
    \includegraphics[width=1.0\linewidth]{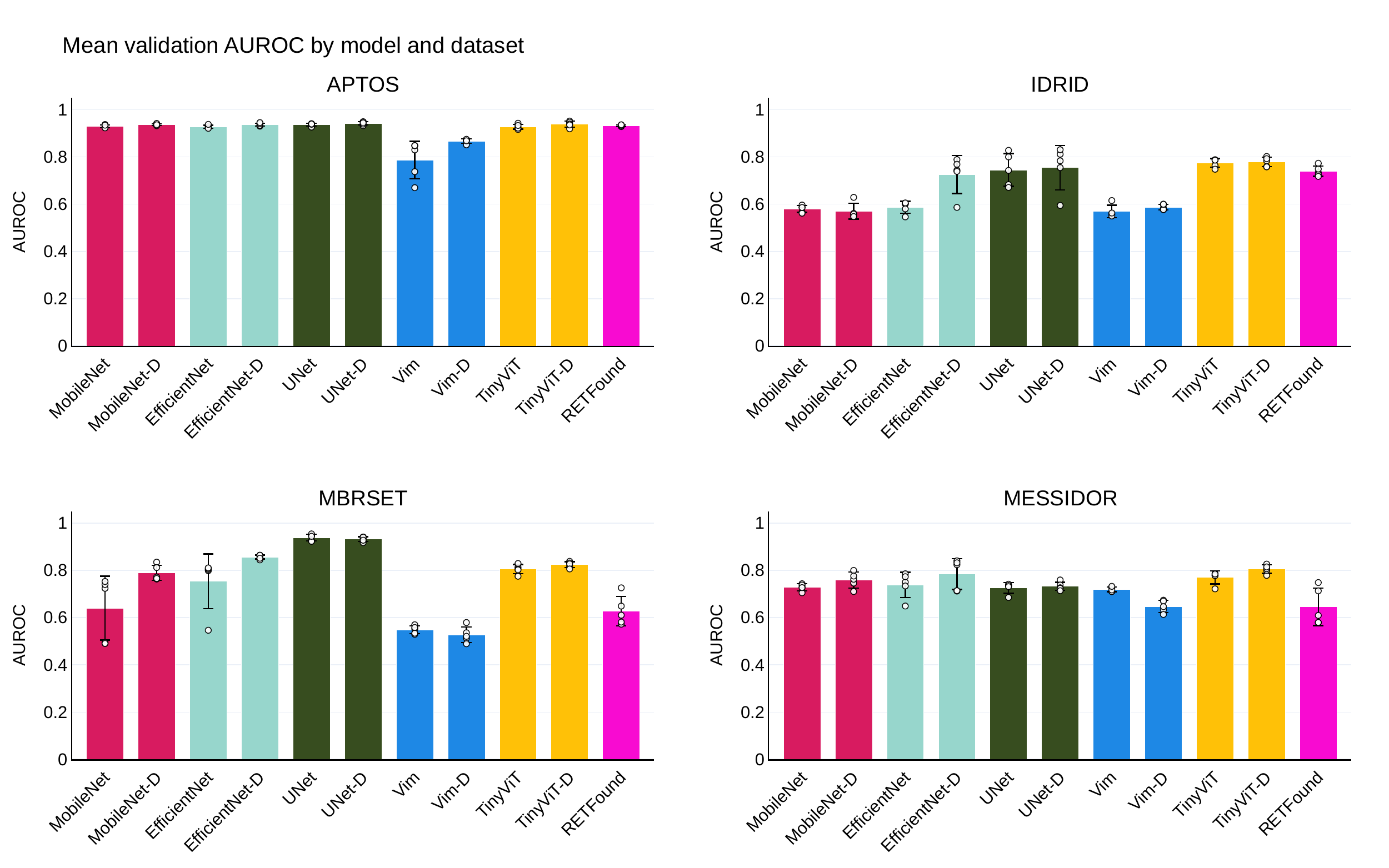}
    \caption{Mean validation AUROC by model and dataset. Bar height shows the mean AUROC, and error bars show the standard deviation. Each dot is the result of a single model run.}
    \label{fig:bar_all}
\end{figure*}

With the masked student established as our primary approach, we present an expanded evaluation against the wider range of competitive baseline architectures. We name each distilled version using the CAMEL framework by the backbone name followed by the identifier ``D'': TinyVit-D, Vim-D, MobileNet-D, EfficientNet-D, and UNet-D. Table~\ref{tab:benchmark_results} summarizes performance across IDRiD, APTOS, mBRSET, and MESSIDOR, while Figure~\ref {fig:bar_all} visually shows these results.

On the IDRiD dataset, the CAMEL framework's impact is notably apparent. TinyVit-D achieves the best overall results across F1 ($0.471 \pm 0.019$) and AUPR ($0.477 \pm 0.037$), improving over the baseline TinyVit. Similarly, UNet-D achieves the highest AUROC ($0.782 \pm 0.089$) on the dataset, substantially improving upon the standard UNet. While RETFound shows the advantages of large-scale foundation pretraining with a slightly better accuracy, the CAMEL-based ViT shows that our knowledge distillation successfully imbues lightweight architectures with diagnostic capabilities, surpassing the foundation model on all other metrics.

On the APTOS dataset, which generally yields higher baseline scores due to more uniform image quality, the CAMEL framework's advantage becomes even more pronounced. Every single distilled variant, MobileNet-D, EfficientNet-D, UNet-D, and TinyVit-D yields strict improvements over its respective baseline in nearly all metrics. Notably, TinyVit-D reaches the highest Accuracy ($0.849 \pm 0.008$) and F1 ($0.736 \pm 0.013$), even surpassing the RETFound teacher. Vim-D also shows surprisingly large gains over the base Vim architecture, raising AUROC from $0.787$ to $0.867$, showing that the distilled backbone retains strong diagnostic discriminability if compared to training from scratch.

On the highly imbalanced mBRSET dataset, real-world handheld device image acquisition, EfficientNet-D shows a remarkable AUROC jump to $0.874 \pm 0.008$ compared to the standard EfficientNet's $0.754 \pm 0.116$. While the standard UNet baseline performs exceptionally well on this specific distribution, TinyVit-D continues to show consistent, broad upgrades over TinyVit across all metrics. Furthermore, Vim-D generalizes well in this challenging environment, outperforming the raw Vim baseline in F1 and showing robustness comparable to standard convolutional networks, yet within a much more structurally efficient framework.

Finally, on the MESSIDOR dataset, TinyVit-D and EfficientNet-D capture the highest scores across F1, AUROC, and AUPR, improving upon their standard versions. Furthermore, Vim-D achieves an accuracy peak ($0.849 \pm 0.003$), outperforming heavier conventional architectures like EfficientNet and MobileNet.

This systematic uplift across diverse architectures confirms that the spatial mapping mechanisms and structural regularization enforced by the CAMEL framework's masking and distillation process may be well-suited to identifying specific lesion structures characteristic of diabetic retinopathy, regardless of the base model topology.

\begin{table}[!htb]
\centering
\caption{Comprehensive Benchmark Performance on validation across IDRiD, APTOS, mBRSET, and MESSIDOR datasets (mean $\pm$ std). The best results in each column per dataset are \textbf{bolded}, and the second best are \underline{underlined}.}
\label{tab:benchmark_results}
\vspace{0.5em}
\resizebox{\linewidth}{!}{%
\begin{tabular}{lcccc}
\toprule
Model & Acc & F1 & AUROC & AUPR \\
\midrule
\multicolumn{5}{c}{\textbf{IDRID}} \\
Vim & 0.254 $\pm$ 0.056 & 0.203 $\pm$ 0.027 & 0.569 $\pm$ 0.026 & 0.296 $\pm$ 0.021 \\
MobileNet & 0.208 $\pm$ 0.059 & 0.168 $\pm$ 0.055 & 0.580 $\pm$ 0.015 & 0.317 $\pm$ 0.031 \\
MobileNet-D & 0.204 $\pm$ 0.027 & 0.190 $\pm$ 0.026 & 0.623 $\pm$ 0.035 & 0.347 $\pm$ 0.036 \\
EfficientNet & 0.235 $\pm$ 0.033 & 0.189 $\pm$ 0.040 & 0.587 $\pm$ 0.025 & 0.304 $\pm$ 0.009 \\
EfficientNet-D & 0.365 $\pm$ 0.037 & 0.322 $\pm$ 0.059 & 0.750 $\pm$ 0.065 & 0.421 $\pm$ 0.053 \\
UNet & 0.361 $\pm$ 0.123 & 0.334 $\pm$ 0.115 & 0.745 $\pm$ 0.069 & 0.406 $\pm$ 0.081 \\
UNet-D & 0.381 $\pm$ 0.121 & 0.359 $\pm$ 0.129 & \textbf{0.782 $\pm$ 0.089} & \textbf{0.493 $\pm$ 0.118} \\
Vim-D & 0.301 $\pm$ 0.028 & 0.255 $\pm$ 0.010 & 0.588 $\pm$ 0.012 & 0.314 $\pm$ 0.006 \\
TinyVit & 0.373 $\pm$ 0.064 & 0.348 $\pm$ 0.058 & 0.775 $\pm$ 0.018 & 0.435 $\pm$ 0.026 \\
TinyVit-D & \underline{0.538 $\pm$ 0.049} & \textbf{0.471 $\pm$ 0.019} & \underline{0.779 $\pm$ 0.020} & \textbf{0.477 $\pm$ 0.037} \\
RETFound & \textbf{0.557 $\pm$ 0.052} & \underline{0.440 $\pm$ 0.034} & 0.740 $\pm$ 0.022 & \underline{0.448 $\pm$ 0.028} \\
\midrule
\multicolumn{5}{c}{\textbf{APTOS}} \\
MobileNet & 0.803 $\pm$ 0.004 & 0.647 $\pm$ 0.017 & 0.930 $\pm$ 0.006 & 0.673 $\pm$ 0.027 \\
MobileNet-D & 0.810 $\pm$ 0.009 & 0.652 $\pm$ 0.027 & \underline{0.940 $\pm$ 0.002} & 0.703 $\pm$ 0.008 \\
EfficientNet & 0.786 $\pm$ 0.019 & 0.619 $\pm$ 0.024 & 0.928 $\pm$ 0.006 & 0.659 $\pm$ 0.020 \\
EfficientNet-D & 0.822 $\pm$ 0.015 & 0.658 $\pm$ 0.037 & 0.941 $\pm$ 0.003 & 0.697 $\pm$ 0.014 \\
UNet & 0.805 $\pm$ 0.019 & 0.644 $\pm$ 0.037 & 0.937 $\pm$ 0.006 & \underline{0.708 $\pm$ 0.022} \\
UNet-D & \underline{0.845 $\pm$ 0.009} & 0.690 $\pm$ 0.030 & \textbf{0.953 $\pm$ 0.002} & \textbf{0.766 $\pm$ 0.013} \\
Vim & 0.588 $\pm$ 0.151 & 0.296 $\pm$ 0.102 & 0.787 $\pm$ 0.080 & 0.430 $\pm$ 0.060 \\
Vim-D & 0.744 $\pm$ 0.004 & 0.445 $\pm$ 0.007 & 0.867 $\pm$ 0.009 & 0.527 $\pm$ 0.014 \\
TinyVit & 0.789 $\pm$ 0.023 & 0.619 $\pm$ 0.048 & 0.928 $\pm$ 0.010 & 0.663 $\pm$ 0.041 \\
TinyVit-D & \textbf{0.849 $\pm$ 0.008} & \textbf{0.736 $\pm$ 0.013} & {0.939 $\pm$ 0.013} & \textbf{0.738 $\pm$ 0.030} \\
RETFound & \underline{0.834 $\pm$ 0.005} & \underline{0.703 $\pm$ 0.010} & 0.932 $\pm$ 0.003 & 0.703 $\pm$ 0.010 \\
\midrule
\multicolumn{5}{c}{\textbf{MBRSET}} \\
MobileNet & 0.799 $\pm$ 0.024 & 0.361 $\pm$ 0.168 & 0.641 $\pm$ 0.135 & 0.369 $\pm$ 0.149 \\
MobileNet-D & 0.803 $\pm$ 0.013 & 0.395 $\pm$ 0.072 & 0.814 $\pm$ 0.016 & 0.482 $\pm$ 0.044 \\
EfficientNet & 0.810 $\pm$ 0.015 & 0.471 $\pm$ 0.166 & 0.754 $\pm$ 0.116 & 0.483 $\pm$ 0.153 \\
EfficientNet-D & 0.834 $\pm$ 0.012 & 0.497 $\pm$ 0.028 & 0.874 $\pm$ 0.008 & 0.580 $\pm$ 0.011 \\
UNet & \textbf{0.895 $\pm$ 0.006} & \textbf{0.693 $\pm$ 0.027} & \underline{0.939 $\pm$ 0.014} & \textbf{0.758 $\pm$ 0.014} \\
UNet-D & \underline{0.879 $\pm$ 0.008} & \underline{0.575 $\pm$ 0.040} & \textbf{0.951 $\pm$ 0.007} & \underline{0.699 $\pm$ 0.055} \\
Vim & 0.759 $\pm$ 0.052 & 0.177 $\pm$ 0.003 & 0.549 $\pm$ 0.017 & 0.226 $\pm$ 0.010 \\
Vim-D & 0.741 $\pm$ 0.058 & 0.180 $\pm$ 0.007 & 0.528 $\pm$ 0.033 & 0.215 $\pm$ 0.014 \\
TinyVit & 0.809 $\pm$ 0.016 & 0.476 $\pm$ 0.034 & 0.805 $\pm$ 0.020 & 0.536 $\pm$ 0.023 \\
TinyVit-D & \underline{0.821 $\pm$ 0.015} & \underline{0.570 $\pm$ 0.022} & {0.824 $\pm$ 0.013} & \underline{0.569 $\pm$ 0.015} \\
RETFound & 0.682 $\pm$ 0.096 & 0.272 $\pm$ 0.008 & 0.628 $\pm$ 0.063 & 0.275 $\pm$ 0.031 \\
\midrule
\multicolumn{5}{c}{\textbf{MESSIDOR}} \\
MobileNet & 0.490 $\pm$ 0.036 & 0.331 $\pm$ 0.031 & 0.728 $\pm$ 0.015 & 0.347 $\pm$ 0.026 \\
MobileNet-D & 0.479 $\pm$ 0.048 & 0.321 $\pm$ 0.065 & 0.777 $\pm$ 0.029 & 0.434 $\pm$ 0.031 \\
EfficientNet & 0.536 $\pm$ 0.023 & 0.405 $\pm$ 0.113 & 0.738 $\pm$ 0.054 & 0.412 $\pm$ 0.099 \\
EfficientNet-D & 0.559 $\pm$ 0.035 & \underline{0.461 $\pm$ 0.112} & \underline{0.785 $\pm$ 0.063} & \underline{0.485 $\pm$ 0.118} \\
UNet & 0.466 $\pm$ 0.042 & 0.315 $\pm$ 0.020 & 0.726 $\pm$ 0.023 & 0.331 $\pm$ 0.021 \\
UNet-D & 0.478 $\pm$ 0.023 & 0.262 $\pm$ 0.021 & 0.754 $\pm$ 0.006 & 0.371 $\pm$ 0.012 \\
Vim & 0.441 $\pm$ 0.047 & 0.322 $\pm$ 0.030 & 0.719 $\pm$ 0.010 & 0.321 $\pm$ 0.020 \\
Vim-D & \textbf{0.849 $\pm$ 0.003} & 0.184 $\pm$ 0.000 & 0.647 $\pm$ 0.025 & 0.272 $\pm$ 0.048 \\
TinyVit & 0.526 $\pm$ 0.022 & 0.409 $\pm$ 0.034 & 0.770 $\pm$ 0.027 & 0.433 $\pm$ 0.063 \\
TinyVit-D & \underline{0.571 $\pm$ 0.020} & \textbf{0.552 $\pm$ 0.028} & \textbf{0.805 $\pm$ 0.018} & \textbf{0.532 $\pm$ 0.017} \\
RETFound & 0.523 $\pm$ 0.042 & 0.299 $\pm$ 0.016 & 0.646 $\pm$ 0.080 & 0.358 $\pm$ 0.076 \\
\bottomrule
\end{tabular}
}
\label{table:results}
\end{table}

\begin{figure}[!htb]
    \centering
    \includegraphics[width=0.9\linewidth]{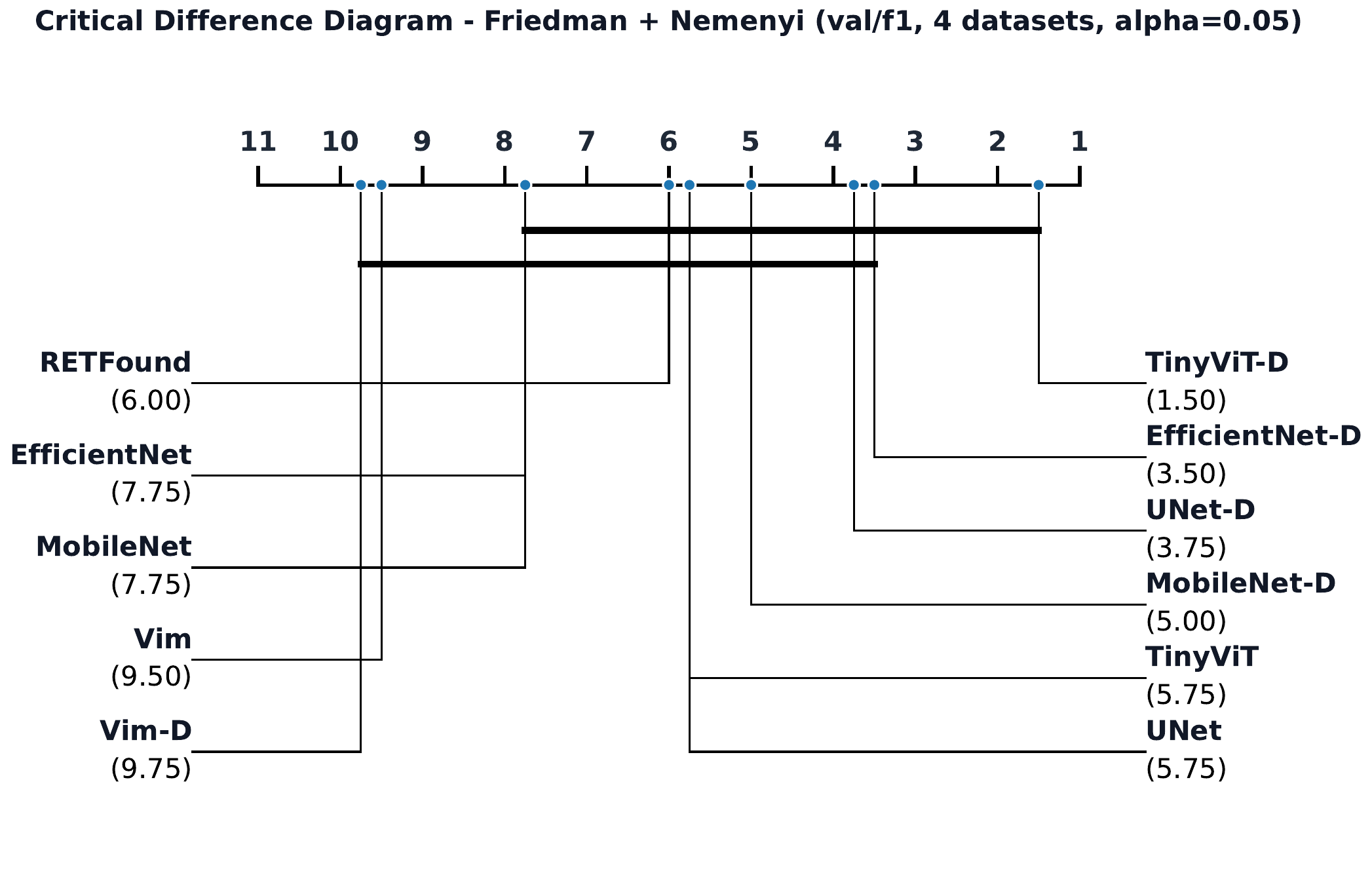}
    \caption{Critical difference (CD) diagram for AUROC. Models connected by a horizontal bar are not significantly different according to the Friedman test with Nemenyi post-hoc correction ($\alpha = 0.05$).}
    \label{fig:cd_diagram}
\end{figure}

The core justification for CAMEL is that its masking mechanism acts as a strong regularizer while balancing classification power with strict constraints on parameters, FLOPs, and inference time. While massive models like RETFound or heavier CNN architectures maximize diagnostic scores, they are fundamentally unsuited for mobile, resource-limited deployments. By asymmetrically distilling RETFound's robust capabilities into a compact masked model, CAMEL minimizes the classification decay often associated with lightweight models. The statistical parity observed (Figure~\ref{fig:cd_diagram}) further shows that computational reduction does not cause a statistically meaningful loss of diagnostic power, positioning CAMEL as a highly capable, edge-deployable alternative for diabetic retinopathy screening.

\section{Conclusion}\label{sec:conclusion}

In this work, we addressed a critical barrier to deploying ophthalmic machine learning models in decentralized clinical settings: the prohibitive computational cost of Transformer-based foundation models. To bridge this gap, we introduced CAMEL. This generalizable feature-distillation framework transfers the rich ophthalmic representations of massive foundation models (such as the 303M-parameter RETFound) into compact student backbones across heterogeneous architectural paradigms. Our findings strongly advocate for knowledge distillation as an essential strategy in this scenario, confirming that the representational power required for accurate retinal screening can be condensed into lightweight networks suitable for edge and mobile deployment without compromising diagnostic precision.

Across extensive benchmarking on four diverse datasets (IDRiD, APTOS, mBRSET, and MESSIDOR-2), CAMEL-distilled models consistently outperformed their counterparts trained from scratch. Crucially, specific distilled student variants, such as TinyViT-D and EfficientNet-D, not only closed the performance gap with the teacher but also surpassed RETFound across key diagnostic metrics, while reducing parameter counts by up to $75\times$ and floating operations by up to $155\times$. By evaluating a broad spectrum of CNN, ViT, and State-Space Model backbones, CAMEL establishes an effective Pareto frontier of efficiency-versus-accuracy trade-offs. This enables practitioners and deployment engineers to select the optimal operational point tailored to their specific hardware, latency, and clinical constraints. Furthermore, our analysis confirms that asymmetric masked feature distillation provides crucial structural regularization, significantly boosting robustness under severe domain and acquisition shifts.

Future work will extend this framework to optimize cross-architecture feature alignment further. We plan to investigate hybrid distillation strategies that combine feature-level embedding alignment with logit-based loss and masked reconstruction objectives to make accurate task-specific models. Additionally, we aim to scale the distillation process using larger, multi-center repositories, such as PAPILA~\cite{kovalyk2022papila} and MESSIDOR~\cite{Decenciere2014messidor}, to assess robustness across broader clinical demographics. Beyond diabetic retinopathy grading, we will evaluate the CAMEL framework's generality across a wider spectrum of ophthalmic tasks, including glaucoma detection, retinal vessel segmentation, and microaneurysm identification.

\bibliographystyle{ACM-Reference-Format}
\bibliography{references}

@article{sevgi2024foundation,
  title={Foundation models in ophthalmology: opportunities and challenges},
  author={Sevgi, Mertcan and Ruffell, Eden and Antaki, Fares and Chia, Mark A and Keane, Pearse A},
  journal={Current opinion in ophthalmology},
  volume={36},
  number={1},
  pages={90},
  year={2024}
}

@article{TING2019100759,
title = {Deep learning in ophthalmology: The technical and clinical considerations},
journal = {Progress in Retinal and Eye Research},
volume = {72},
pages = {100759},
year = {2019},
issn = {1350-9462},
doi = {https://doi.org/10.1016/j.preteyeres.2019.04.003},
url = {https://www.sciencedirect.com/science/article/pii/S1350946218300909},
author = {Daniel S.W. Ting and Lily Peng and Avinash V. Varadarajan and Pearse A. Keane and Philippe M. Burlina and Michael F. Chiang and Leopold Schmetterer and Louis R. Pasquale and Neil M. Bressler and Dale R. Webster and Michael Abramoff and Tien Y. Wong}
}

@article{PIZER1987355,
title = {Adaptive histogram equalization and its variations},
journal = {Computer Vision, Graphics, and Image Processing},
volume = {39},
number = {3},
pages = {355-368},
year = {1987},
issn = {0734-189X},
doi = {https://doi.org/10.1016/S0734-189X(87)80186-X},
url = {https://www.sciencedirect.com/science/article/pii/S0734189X8780186X},
author = {Stephen M. Pizer and E. Philip Amburn and John D. Austin and Robert Cromartie and Ari Geselowitz and Trey Greer and Bart {ter Haar Romeny} and John B. Zimmerman and Karel Zuiderveld}
}

@INPROCEEDINGS{9150790,
  author={Cubuk, Ekin D. and Zoph, Barret and Shlens, Jonathon and Le, Quoc V.},
  booktitle={2020 IEEE/CVF Conference on Computer Vision and Pattern Recognition Workshops (CVPRW)}, 
  title={Randaugment: Practical automated data augmentation with a reduced search space}, 
  year={2020},
  volume={},
  number={},
  pages={3008-3017},
  doi={10.1109/CVPRW50498.2020.00359}}

@article{zhou2023retfound,
	title = {A foundation model for generalizable disease detection from retinal images ({RetFound})},
	volume = {622},
	copyright = {https://creativecommons.org/licenses/by/4.0},
	issn = {0028-0836, 1476-4687},
	url = {https://www.nature.com/articles/s41586-023-06555-x},
	doi = {10.1038/s41586-023-06555-x},
	language = {en},
	number = {7981},
	urldate = {2025-05-22},
	journal = {Nature},
	author = {Y. Zhou and Chia, Mark A. and Wagner, Siegfried K. and Ayhan, Murat S. and Williamson, Dominic J. and Struyven, Robbert R. and Liu, Timing and Xu, Moucheng and Lozano, Mateo G. and Woodward-Court, Peter and Kihara, Yuka and {UK Biobank Eye \& Vision Consortium} and Allen, Naomi and Gallacher, John E. J. and Littlejohns, Thomas and Aslam, Tariq and Bishop, Paul and Black, Graeme and Sergouniotis, Panagiotis and Atan, Denize and Dick, Andrew D. and Williams, Cathy and Barman, Sarah and Barrett, Jenny H. and Mackie, Sarah and Braithwaite, Tasanee and Carare, Roxana O. and Ennis, Sarah and Gibson, Jane and Lotery, Andrew J. and Self, Jay and Chakravarthy, Usha and Hogg, Ruth E. and Paterson, Euan and Woodside, Jayne and Peto, Tunde and Mckay, Gareth and Mcguinness, Bernadette and Foster, Paul J. and Balaskas, Konstantinos and Khawaja, Anthony P. and Pontikos, Nikolas and Rahi, Jugnoo S. and Lascaratos, Gerassimos and Patel, Praveen J. and Chan, Michelle and Chua, Sharon Y. L. and Day, Alexander and Desai, Parul and Egan, Cathy and Fruttiger, Marcus and Garway-Heath, David F. and Hardcastle, Alison and Khaw, Sir Peng T. and Moore, Tony and Sivaprasad, Sobha and Strouthidis, Nicholas and Thomas, Dhanes and Tufail, Adnan and Viswanathan, Ananth C. and Dhillon, Bal and Macgillivray, Tom and Sudlow, Cathie and Vitart, Veronique and Doney, Alexander and Trucco, Emanuele and Guggeinheim, Jeremy A. and Morgan, James E. and Hammond, Chris J. and Williams, Katie and Hysi, Pirro and Harding, Simon P. and Zheng, Yalin and Luben, Robert and Luthert, Phil and Sun, Zihan and McKibbin, Martin and O’Sullivan, Eoin and Oram, Richard and Weedon, Mike and Owen, Chris G. and Rudnicka, Alicja R. and Sattar, Naveed and Steel, David and Stratton, Irene and Tapp, Robyn and Yates, Max M. and Petzold, Axel and Madhusudhan, Savita and Altmann, Andre and Lee, Aaron Y. and Topol, Eric J. and Denniston, Alastair K. and Alexander, Daniel C. and Keane, Pearse A.},
	month = oct,
	year = {2023},
	note = {Publisher: Springer Science and Business Media LLC},
	pages = {156--163},
}

@misc{zhu2024Vim,
      title={Vision Mamba: Efficient Visual Representation Learning with Bidirectional State Space Model}, 
      author={Lianghui Zhu and Bencheng Liao and Qian Zhang and Xinlong Wang and Wenyu Liu and Xinggang Wang},
      year={2024},
      eprint={2401.09417},
      archivePrefix={arXiv},
      primaryClass={cs.CV},
      url={https://arxiv.org/abs/2401.09417}, 
}

@misc{shi2024eyefound,
      title={EyeFound: A Multimodal Generalist Foundation Model for Ophthalmic Imaging}, 
      author={Danli Shi and Weiyi Zhang and Xiaolan Chen and Yexin Liu and Jiancheng Yang and Siyu Huang and Yih Chung Tham and Yingfeng Zheng and Mingguang He},
      year={2024},
      eprint={2405.11338},
      archivePrefix={arXiv},
      primaryClass={cs.CV},
      url={https://arxiv.org/abs/2405.11338}, 
}

@InProceedings{du2024retclip,
author="Du, Jiawei
and Guo, Jia
and Zhang, Weihang
and Yang, Shengzhu
and Liu, Hanruo
and Li, Huiqi
and Wang, Ningli",
editor="Linguraru, Marius George
and Dou, Qi
and Feragen, Aasa
and Giannarou, Stamatia
and Glocker, Ben
and Lekadir, Karim
and Schnabel, Julia A.",
title="RET-CLIP: A Retinal Image Foundation Model Pre-trained with Clinical Diagnostic Reports",
booktitle="Medical Image Computing and Computer Assisted Intervention -- MICCAI 2024",
year="2024",
publisher="Springer Nature Switzerland",
address="Cham",
pages="709--719",
isbn="978-3-031-72390-2"
}

@article{caron2021dino,
  title={Emerging properties in self-supervised vision transformers},
  author={Caron, Mathilde and Touvron, Hugo and Misra, Ishan and J{\'e}gou, Herv{\'e} and Mairal, Julien and Bojanowski, Piotr and Joulin, Armand},
  booktitle={2021 IEEE/CVF international conference on computer vision (ICCV)},
  pages={9630--9640},
  year={2021},
  organization={IEEE},
  journal      = {CoRR},
  volume       = {abs/2104.14294},
  bibsource    = {dblp computer science bibliography, https://dblp.org}
}

@article{porwal2018idrid,
  title={Indian diabetic retinopathy image dataset (IDRiD): a database for diabetic retinopathy screening research},
  author={Porwal, Prasanna and Pachade, Samiksha and Kamble, Ravi and Kokare, Manesh and Deshmukh, Girish and Sahasrabuddhe, Vivek and Meriaudeau, Fabrice},
  journal={Data},
  volume={3},
  number={3},
  pages={25},
  year={2018},
  publisher={MDPI}
}

@INPROCEEDINGS{kassani2019aptos,
  author={Kassani, Sara Hosseinzadeh and Kassani, Peyman Hosseinzadeh and Khazaeinezhad, Reza and Wesolowski, Michal J. and Schneider, Kevin A. and Deters, Ralph},
  booktitle={2019 IEEE International Symposium on Signal Processing and Information Technology (ISSPIT)}, 
  title={Diabetic Retinopathy Classification Using a Modified Xception Architecture}, 
  year={2019},
  pages={1-6},
publisher={IEEE},
address={New York, NY, USA},
  doi={10.1109/ISSPIT47144.2019.9001846}
}

@article{kovalyk2022papila,
	title = {{PAPILA}: {Dataset} with fundus images and clinical data of both eyes of the same patient for glaucoma assessment},
	volume = {9},
	issn = {2052-4463},
	url = {https://doi.org/10.1038/s41597-022-01388-1},
	doi = {10.1038/s41597-022-01388-1},
	number = {1},
	journal = {Scientific Data},
	author = {Kovalyk, Oleksandr and Morales-Sánchez, Juan and Verdú-Monedero, Rafael and Sellés-Navarro, Inmaculada and Palazón-Cabanes, Ana and Sancho-Gómez, José-Luis},
	month = jun,
	year = {2022},
	pages = {291},
}

@article{Decenciere2014messidor,
  title={Feedback on a publicly distributed image database: the Messidor database},
  author={Decenci{\`e}re, Etienne and Zhang, Xiwei and Cazuguel, Guy and Lay, Bruno and Cochener, B{\'e}atrice and Trone, Caroline and Gain, Philippe and Ord{\'o}{\~n}ez-Varela, John-Richard and Massin, Pascale and Erginay, Ali and others},
  journal={Image Analysis \& Stereology},
  volume={33},
  number={3},
  pages={231--234},
  year={2014}
}

@misc{tinyvit,
      title={TinyViT: Fast Pretraining Distillation for Small Vision Transformers}, 
      author={Kan Wu and Jinnian Zhang and Houwen Peng and Mengchen Liu and Bin Xiao and Jianlong Fu and Lu Yuan},
      year={2022},
      eprint={2207.10666},
      archivePrefix={arXiv},
      primaryClass={cs.CV},
      url={https://arxiv.org/abs/2207.10666}, 
}

@inproceedings{EfficientNet,
  title={EfficientNet: Rethinking Model Scaling for Convolutional Neural Networks},
  author={Tan, Mingxing and Le, Quoc V.},
  booktitle={Proceedings of the 36th International Conference on Machine Learning (ICML)},
  pages={6105--6114},
  year={2019},
publisher={PMLR},
  address={Long Beach, CA, USA}
}

@inproceedings{mobilenet,
  title={Searching for MobileNetV3},
  author={Howard, Andrew and Sandler, Mark and Chu, Grace and Chen, Liang-Chieh and Chen, Bo and Tan, Mingxing and Wang, Weijun and Zhu, Yukun and Pang, Ruoming and Vasudevan, Vijay and Le, Quoc V. and Adam, Hartwig},
  booktitle={Proceedings of the IEEE/CVF International Conference on Computer Vision (ICCV)},
  pages={1314--1324},
  year={2019},
publisher={IEEE},
  address={New York, NY, USA}
}

@inproceedings{unet,
  title={U-net: Convolutional networks for biomedical image segmentation},
  author={Ronneberger, Olaf and Fischer, Philipp and Brox, Thomas},
  booktitle={International Conference on Medical Image Computing and Computer-Assisted Intervention (MICCAI)},
  pages={234--241},
  year={2015},
  organization={Springer},
address={Cham, Switzerland},
publisher={Springer International Publishing},
}

@misc{mbrset,
  author = {Nakayama, Luis Filipe and {Zago Ribeiro}, Lucas and Restrepo, David and {Santos Barboza}, Nathan and {Dias Fiterman}, Raul and {Vieira Sousa}, Maria luiza and Pereira, Alexandre Durao Alves and Regatieri, Caio and Malerbi, Fernando Korn and Andrade, Rafael},
  title = {{mBRSET, a Mobile Brazilian Retinal Dataset}},
  journal = {{PhysioNet}},
  year = {2024},
  month = jun,
  note = {Version 1.0},
  doi = {10.13026/qxpd-1y65},
  url = {https://doi.org/10.13026/qxpd-1y65},
    numpages = {11},
    howpublished = {{PhysioNet}},
}

@inproceedings{he2022masked,
  title={Masked autoencoders are scalable vision learners},
  author={He, Kaiming and Chen, Xinlei and Xie, Saining and Li, Yanghao and Doll{\'a}r, Piotr and Girshick, Ross},
  booktitle={Proceedings of the IEEE/CVF Conference on Computer Vision and Pattern Recognition (CVPR)},
  pages={15979--15988},
publisher={IEEE},
  year={2022},
  organization={IEEE},
address={New York, NY, USA}
}

@misc{yang2022maskedgenerativedistillation,
      title={Masked Generative Distillation}, 
      author={Zhendong Yang and Zhe Li and Mingqi Shao and Dachuan Shi and Zehuan Yuan and Chun Yuan},
      year={2022},
      eprint={2205.01529},
      archivePrefix={arXiv},
      primaryClass={cs.CV},
      url={https://arxiv.org/abs/2205.01529}, 
}

@article{hinton2015distillation,
  author  = {Hinton, Geoffrey and Vinyals, Oriol and Dean, Jeff},
  title   = {Distilling the Knowledge in a Neural Network},
  journal = {CoRR},
  volume  = {abs/1503.02531},
  year    = {2015},
  url     = {https://arxiv.org/abs/1503.02531},
    numpages = {9}
}

@misc{heidari2024efficiency,
      title={Enhancing Efficiency in Vision Transformer Networks: Design Techniques and Insights}, 
      author={Moein Heidari and Reza Azad and Sina Ghorbani Kolahi and René Arimond and Leon Niggemeier and Alaa Sulaiman and Afshin Bozorgpour and Ehsan Khodapanah Aghdam and Amirhossein Kazerouni and Ilker Hacihaliloglu and Dorit Merhof},
      year={2024},
      eprint={2403.19882},
      archivePrefix={arXiv},
      primaryClass={eess.IV},
      url={https://arxiv.org/abs/2403.19882}, 
}

@article{bommasani2022,
  title={On the Opportunities and Risks of Foundation Models},
  author={Bommasani, Rishi and others},
  journal={arXiv preprint arXiv:2108.07258},
  year={2022},
number={1},
  pages={1--212},
volume={abs/2108.07258},
  url={https://arxiv.org/abs/2108.07258}
}


\end{document}